\documentclass[conference]{IEEEtran}
\IEEEoverridecommandlockouts
\usepackage{cite}
\usepackage{amsmath,amssymb,amsfonts}
\usepackage{algorithmic}
\usepackage{graphicx}
\usepackage{siunitx}
\usepackage{booktabs}
\usepackage{textcomp}
\usepackage{xcolor}
\usepackage[ruled]{algorithm2e}
\usepackage{bbding}
\usepackage{pifont}
\usepackage{fancyhdr}
\usepackage{soul}
\soulregister\ref7
\soulregister\cite7 
\soulregister\citep7 
\soulregister\citet7 

\usepackage{indentfirst}
\usepackage[ruled]{algorithm2e}
\def\BibTeX{{\rm B\kern-.05em{\sc i\kern-.025em b}\kern-.08em
    T\kern-.1667em\lower.7ex\hbox{E}\kern-.125emX}}

\begin{document}

\title{
Enabling Efficient Deep Convolutional Neural Network-based Sensor Fusion for Autonomous Driving
}


\author{\IEEEauthorblockN{Xiaoming Zeng\normalfont\textsuperscript{*}, Zhendong Wang\normalfont\textsuperscript{*}, 
Yang Hu, {\{xiaoming.zeng, zhendong.wang, yang.hu4\}@utdallas.edu}
}
\IEEEauthorblockA{\textit{Electrical and Computer Engineering Department, The University of Texas at Dallas, Richardson, US}
}
}

\maketitle
\footnotetext[1]{These two authors contributed equally.}

\begin{abstract}







Autonomous driving demands accurate perception and safe decision-making. To achieve this, automated vehicles are now equipped with multiple sensors (e.g., camera, Lidar, etc.), enabling them to exploit complementary environmental context by fusing data from different sensing modalities. With the success of Deep Convolutional Neural Network(DCNN), the fusion between DCNNs has been proved as a promising strategy to achieve satisfactory perception accuracy. However, mainstream existing DCNN fusion schemes conduct fusion by directly element-wisely adding feature maps extracted from different modalities together at various stages, failing to consider whether the features being fused are matched or not. Therefore, we first propose a feature disparity metric to quantitatively measure the degree of feature disparity between the feature maps being fused. We then propose Fusion-filter as a feature-matching techniques to tackle the feature-mismatching issue.  We also propose a Layer-sharing technique in the deep layer that can achieve better accuracy with less computational overhead. Together with the help of the feature disparity to be an additional loss, our proposed technologies enable DCNN to learn corresponding feature maps with similar characteristics and complementary visual context from different modalities to achieve better accuracy. Experimental results demonstrate that our proposed fusion technique can achieve better accuracy on KITTI dataset with less computational resources demand.
\end{abstract}


\section{Introduction}
\begin{figure}[t]
\centering
            \includegraphics[width=0.6\linewidth]{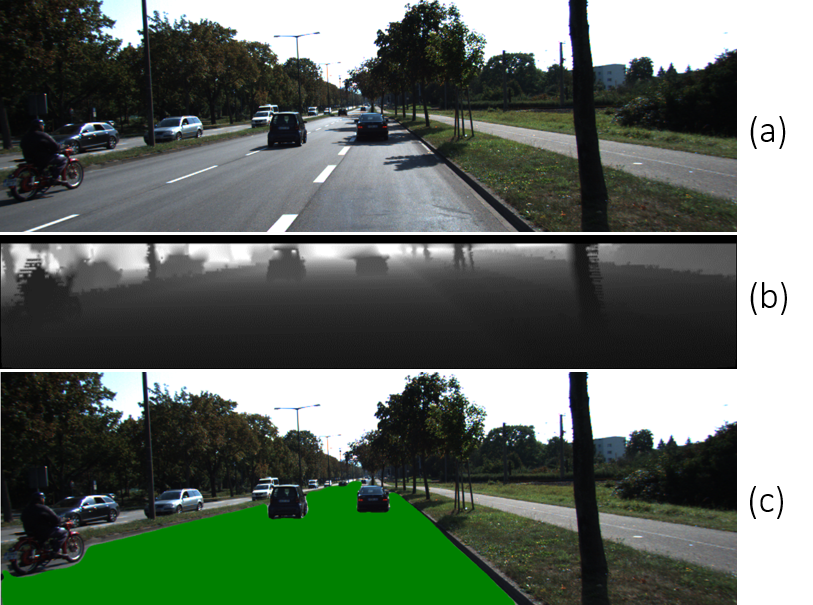}
            \vspace{-3mm}
            \caption{The free road segmentation result by fusion.  (a) RGB input image. (b) Depth input image. (c) segmentation result. The free (drivable) road is represented in green pixels.}\label{segresult}
            \vspace{-4mm}
\end{figure}

The era of driving automation is coming. The safety of an automated vehicle hinges crucially upon the accuracy of perception. Therefore, many studies \cite{cross,fusenet,sne, progressive, late1, early1} have employed multi-modal sensors such as cameras and LiDARs that can provide complementary sensing information to deliver better and robust perception performance. In this paper, we focus on the free road segmentation since it is a cornerstone module among all autonomous driving tasks. It is a typical application that benefits from such multi-modal sensing technology as shown in Fig.\ref{segresult}. In this multi-modal sensor fusion setup, both RGB images(captured from cameras) and depth images(pre-processed from 3D point cloud collected by LiDAR) are employed, as depicted in Fig.\ref{segresult} (a) and Fig.\ref{segresult} (b) respectively. We can observe that the RGB and depth images are a pair of interpretations of the same scene at the same moment, providing an opportunity to exploit them for a better perception accuracy. 
In contrast, employing only one sensing modality often fails the task in some driving scenarios, for example, using only RGB camera under unfavorable lighting conditions such as dark night, overexposure, and etc. 

Among these fusion architectures, though there is no conclusive evidence that one fusion method is absolutely better than others, the middle fusion method with element-wise summation \cite{progressive,sne,fusenet,cross} is the dominant method in recent works as we can notice in the KITTI benchmarks \cite{kitti}. The reason is two-fold. First, the middle fusion architecture often adopts crossing connections between network branches in all fusion stages (i.e., from shallow to deep layer), and these crossing connections can learn characteristics from the training data, that where and to what degree the integration should be carried out \cite{cross}, and thus providing better accuracy compared to the early \cite{early1} and late fusions \cite{late1} architectures. Second, fusion operations can be achieved by executing element-wise summation between corresponding intermediate feature maps, which is easy to be implemented.

Although it is a fact that middle fusion with element-wise summation provide the best accuracy to date, we frustratedly observe that existing solutions all fail to consider the \textit{intrinsic relationship between two sets of feature maps to be fused}. Specifically, naively conducting the element-wise summation of intermediate feature maps from multi-modal sensing data can lead to such a scenario, where a sensor A's feature map A\_1 that represents the feature X is fused with a sensor B's feature map B\_1 that represents the feature Y, since A\_1 and B\_1 is product of convolution using different filters.
the fusion of mismatched features can generate meaningless information that might harm the accuracy and eventually threat the driving safety. 

In this work, we set out to explore the impacts of feature matching to the accuracy of sensor fusion. Specifically, we propose a \textbf{\textit{feature disparity (FD)}} metric to quantitatively describe the differences between features from different sensing modals. Consequently, we argue that in middle fusion the feature maps to be fused should possess similar visual characteristic with complementary content, that being said, with low feature disparity. 
To achieve this goal, we first propose \textbf{\textit{Fusion filter}} that learns the feature-matching relationship between the feature maps to be fused to guarantee the feature matching. 
Second, we propose \textbf{\textit{Layer-sharing}} network architecture which allows the deep layers to share the same filters based on our observation that features processed in deep layers tend to be similar 
and the feature matching is preserved. Besides, we utilize the feature disparity metric as an additional sub-objective loss function(i.e., \textbf{\textit{Feature Disparity Loss}}) to further constrain two sub-network branches to learn similar features during the training stage. This will not be executed in the inference stage, hence does not affect the inference latency.

Finally, we implement comprehensive evaluations on models with our proposed techniques on the KITTI dataset \cite{kitti}, and demonstrate that the feature-matching techniques can effectively reduce the feature disparity between the feature maps to be fused and achieve better accuracy than that of the state-of-the-art RoadSeg \cite{sne} adopting the naive fusion. 
Meanwhile, the Layer-sharing technique can effectively reduce the computational overhead of the fused network with a comparable or better accuracy compared to the RoadSeg.

\noindent{To summarize, this paper makes the following contributions:}
\begin{itemize}
    \item To the best of our knowledge, we are the first to identify the feature-mismatching issue while performing element-wise fusion in a DCNN-based middle fusion method. Accordingly, a \textbf{\textit{feature disparity metric}} is proposed to quantitatively measure the degree of feature deviation between feature maps to be fused. Feature disparity is then adopted as a individual loss in addition to the baseline segmentation loss, enabling DCNN to learn similar features with complementary content extracted from different network branches.

    \item We propose the technique, \textbf{\textit{Fuse-filter}}, to address the feature-mismatching issue when fusing two independent DCNN feature maps at various stages. Furthermore, we propose the \textbf{\textit{Layer-sharing}} network architecture which allows the deep layers to share the same filters in the fused network so that the feature matching property is preserved.
    
     \item Our evaluation results among models equipped with different proposed fusion schemes on KITTI road dataset reveal that our proposed \textbf{\textit{Fuse-filter}} can achieve better accuracy, while \textbf{\textit{Layer-sharing}} can obtain comparable accuracy with less computational  resources  demand on  KITTI  dataset.
     
     
  
\end{itemize}


\section{Background and Challenges}
 
\subsection{DCNN-based Sensor Fusion}\label{basemodel}
As a representative subset of DNN architecture, the Deep Convolutional Neural Network(DCNN) is widely apply in vision-related tasks \cite{yolo,imgtask1,imgtask3}. Typically, DCNN is a stacked structure composed of multiple layers \cite{lecun}. From shallow to deep, convolutional layers can hierarchically extract embedded visual features 
by following the sliding-window method from traditional image processing \cite{filterslide}. Specifically, each set of filters slides over the input image and performs convolution operation to produce the output feature map with a certain characteristic corresponding to the content inside the sliding filter, that can perform edge extraction, image sharpening, etc. 

As a mainstream sensor fusion method, the DCNN-based fusion architecture leverages multiple separate neural networks to process the data from the multiple independent sensors as the perception approaches using single-modality data (e.g., either the camera data or the LiDAR data) can fail the task. For instance, the LiDAR point cloud data is sparse and without the fine texture of the object being scanned upon, using only LiDAR data might lead to unfavorable perception accuracy.

\begin{figure}[t]
\centering
            \includegraphics[width=0.8\linewidth]{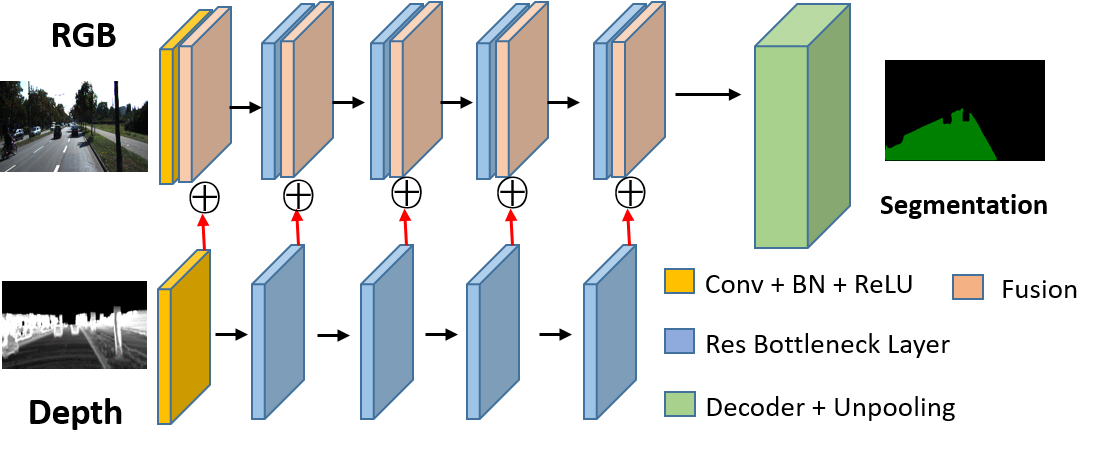}
            \vspace{-3mm}
            \caption{Architecture of baseline RoadSeg.}\label{baseline}
\end{figure}

In this work, We target the \textbf{free road segmentation} task, which is crucial for the driving system to distinguish the free drivable road from the surroundings.  \textbf{RoadSeg} \cite{sne}, as our baseline, is the state-of-the-art open-source DCNN architecture with a middle fusion element-wise summation scheme between the separate RGB and Depth network branch, ranking at the top of KITTI road segmentation task. It also adopts the popular encoder and decoder architecture with ResNet \cite{resnet} being the backbone, which is shown in Fig. \ref{baseline}. Specifically, an RGB encoder and a Depth encoder are employed as the two branches to extract the features from RGB and Depth images, respectively. Meanwhile, at each fusion stage, the extracted RGB and depth features are fused via the element-wise summation operation. Then, the fused feature maps from the encoder are fed to the decoder to generate the final drivable road segmentation result. 

\subsection{Feature Disparity Assessment}\label{conventionalFC}
Feature disparity assessment \cite{survey} is a fundamental process in a variety of traditional computer vision tasks, which is a procedure to compare the feature disparity between two correspondent images.
 The most standard feature disparity measure is L2 metric, which naively compares the pixel-level value difference between two images.

Beyond the standard L2 metric, various methods to measure disparity between features have been proposed in traditional computer vision field \cite{ssim,survey,mi,crossbin} as shown in first two rows of Table.\ref{matchmetric}.
Many feature disparity metrics such as mutual information (MI) \cite{mi} and the cross-bin \cite{crossbin} mainly focus on the statistical pixel-level mean and variation in luminance, lacking of the spatial information embedded in the feature map, which is deemed as the major feature in our task. And though the structural disparity measure (SSIM) \cite{ssim} takes the structural information into account, it  
favors two images to be similar in terms of pixel-level intensity in luminance all across the image, which is not applicable in our case.

\subsection{Problem Definition}\label{Challenges}
As we focus on the DCNN-based middle fusion architecture with the element-wise summation technique, and the fusion is carried out by directly element-wisely adding the intermediate feature maps from different modalities.
Due to the fact that the feature maps to be fused are extracted by the two separate DCNN branches which uses their own associated filters, there are great chances that feature maps to be fused possess mismatched characteristics with each other, simply element-wisely summing the feature maps can cause the chaotic and unfavorable results.  

\begin{table}[t]
\caption{Feature disparity metric comparison}
\begin{tabular}{cccl }
\toprule
Feature disparity metric    &Spatial information &luminance disparity\\
\midrule
MI, Cross-bin   &\XSolid &\XSolid  \\
SSIM  &\Checkmark &\XSolid   \\
Feature Disparity  &\Checkmark &\Checkmark   \\
\bottomrule
\end{tabular}
\label{matchmetric}
\vspace{-4mm}
\end{table}

In order to quantitatively measure the degree of feature deviation between the feature maps to be fused, 
different from the metrics introduced in previous section, our concern is two-fold. First, the spatial features from the two sets of feature maps to be fused should be represented and compared; Second, the pixel-level difference in luminance of two feature maps is expected, since they are obtained from two different sensing modalities. For example, for the same driving scene, the RGB image obtained from camera may be darker during the night, while depth image converted from LiDAR point cloud will not be affected by the light condition. In this case, the overall pixel-level luminance difference would be expected, therefore, previous metrics would fail since they are sensitive to this pixel-level difference in luminance.
Therefore, we choose the edge information as the representative feature of each intermediate feature map, and conduct comparison between the extracted edges. 
Because the edge sketches by nature can well preserve the spatial information and it can also be identified as long as there exists pixel-level difference on different objects despite the overall luminance condition.

By borrowing the edge detection \cite{basu2002gaussian} idea from traditional computer vision field, we adopt opencv edge detection library \cite{opencvedge} to extract the edge sketches of each feature map, then the Feature Disparity will be obtained by conducting comparison between the extracted edge from their corresponding feature maps, which is formally described in Eq.\ref{loss_match}.

\begin{equation}\label{loss_match}
\mathcal{D}_{fd}=\frac{1}{C}\sum_{i=0}^{C} [\mathcal{E}_{i}(f_{Rc}) - \mathcal{E}_{i}(f_{Dc})]^2
\end{equation}
 Note that the channel-wise edge extraction process, denoted as $\mathcal{E}$, will first be performed on feature maps from both RGB branch($f_{Rc}$) and Depth branch($f_{Dc}$), then the pixel-level feature disparity would be obtained across all the corresponding channels. And $C$ represents the total number of channels.
\begin{figure}[t]
\centering
            \includegraphics[width=\linewidth]{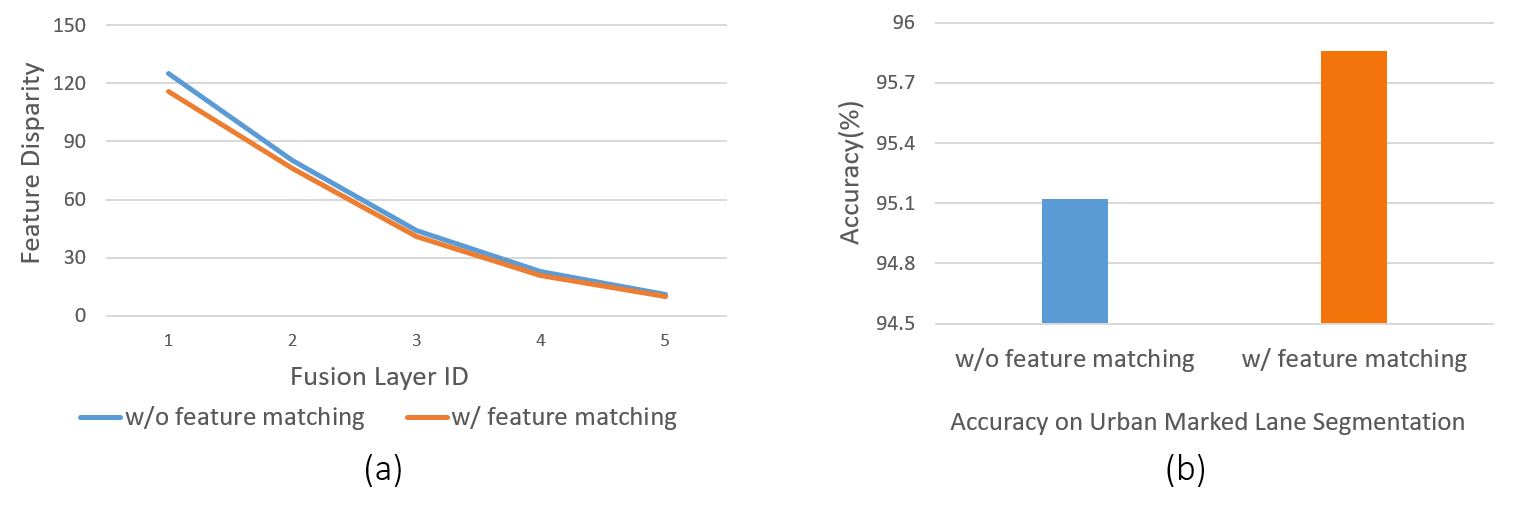}
            \vspace{-3mm}
            \caption{(a)Feature disparity between two sets of feature maps to be fused at different fusion layers.(b)Corresponding accuracy performance evaluated with and without feature-matching technique. }\label{badmatch}
\end{figure}

Fig.\ref{badmatch} (a) shows the result of our proposed feature disparity metric applied on the intermediate RGB and Depth feature maps at five different fusion stages over ten randomly selected RGB and Depth input image pairs. We can see that the feature mismatch between corresponding RGB and Depth feature maps exists at all fusion layers, and it gets less significant as fusion layer gets deeper. The blue line denotes the raw feature disparity in our baseline model, and after applying our proposed feature matching technique in next chapter, the feature disparity can be reduced as the orange line shows.
And a better accuracy performance is also gained accordingly as observed in Fig.\ref{badmatch} (b). 
In addition, we learn that high-level feature maps tend to hold similar feature, offering a opportunity to share the feature-extracting filters residing in deep layers between two network branches. By sharing these deep layers, the model's computational overhead in terms of total number of parameters can be reduced, which will be discussed in detail in next chapter.

\begin{figure}[t]
\centering
            \includegraphics[width=\linewidth]{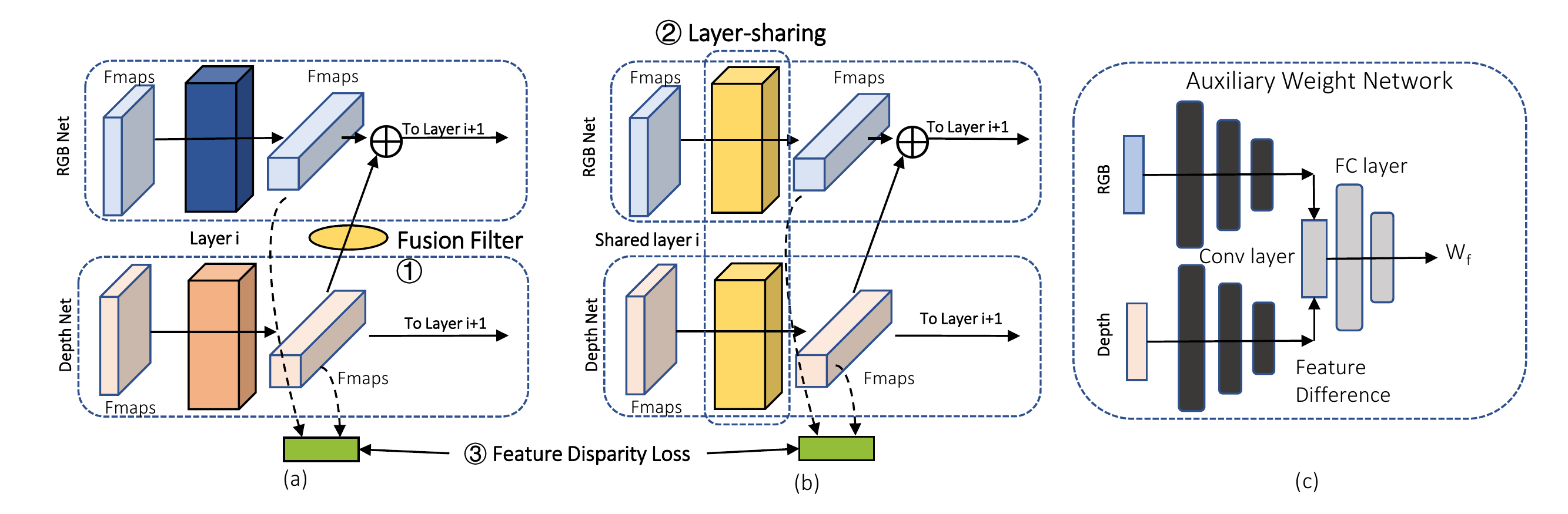}
            \vspace{-5mm}
            \caption{(a) Feature-matching technique: Fusion-filter, which is denoted as a yellow oval box. (b) Layer-sharing method. The shared layer between two network branches is presented as a yellow square. (c) Auxiliary Weight Network. }\label{design}
            \vspace{-2mm}
\end{figure}

\section{Design}
To address the aforementioned feature-mismatching issue, we first present a feature-matching technique by leveraging Fusion-filter to make sure that two feature maps to be fused at each fusion stage are of similar characteristics with complementary content (\ding{192} in Fig.\ref{design} (a)). Meanwhile, we observe that the high-level feature maps from two fusion branches often carry similar characteristic, a Layer-sharing method is proposed to reduce the computational overhead and network parameter volume (\ding{193} in Fig.\ref{design} (b)). 
Furthermore, we apply the feature disparity loss in addition to baseline segmentation loss function to further guide the model to learn filters that can extract analogous features from each modality (\ding{194} in Fig.\ref{design} (a) and (b)).

\subsection{Feature-matching Utilizing Fusion-Filter}\label{AA}

As we have discussed in section \ref{Challenges}, the best fusion paradigm would be those feature maps to be fused hold similar characteristics with complementary content.  
To achieve this goal, we introduce Fusion-filter technique to address the feature-mismatching issue, which 
aims to learn the feature-matching relationship between two sets of feature maps to be fused from training data.

Without the loss of generality, we introduce the Fusion-filter on top of the baseline architecture. As shown in Fig. \ref{design}(a), the Fusion-filter, presented as a yellow circle between two separate neural networks, is employed before where the Depth feature maps are element-wisely summed with RGB feature maps. The usage of Fusion-filter can be described as Eq. \ref{filter}. At the fusion stage $i$, the Depth feature maps $f_{Di}$ are first convoluted with the corresponding fusion filter $W_{f}$.  This convolution process is denoted as $F_{f}$. And the resulting intermediate RGB feature maps $f^{'}_{Ri}$ is updated by the summation between original RGB feature maps $f_{Ri}$ and the aforementioned convolution result. 

\begin{equation}\label{filter}
f^{'}_{R_{i}} = f_{R_{i}} + F_{f}(f_{D_{i}} ; W_{f})
\end{equation}

The Fusion-filter is designed to reconstruct the Depth feature maps by conducting a convolution with Fusion-filter $W_{f}$, which is capable of learning the matching relationship from Depth to RGB feature maps from the training data. Note that the kernel size of the fusion filter is 1x1, since it only aims at reorganizing the mapping relationship between those two sets of feature maps.
On the other hand, the extra memory access for Fusion-filter parameters and corresponding convolution calculation will be introduced, which will be further discussed it in section \ref{eval}.

\subsection{Leveraging Layer-sharing Method}\label{LayersharingMethod}

As Fig. \ref{badmatch} (a) shows, with layer going deeper, the feature disparity between the two sets of feature maps significantly decreases, which implies a great opportunity to propose a Layer-sharing technique that allows the two fused networks to share filters in the deep layers while the shallow layers stay the same. 
Fig. \ref{design} (b) shows an example of sharing convolution layer (indicated by the yellow rectangle) between the two networks. That is, the Depth and RGB network branches share the same convolution filters, while processing their own data independently.


Furthermore, after further analyzing the baseline architecture, we find that in non-shared architecture, there is an implicit weight embedded in each set of filters from two network branches when the fusion is carried out. Therefore, we propose to apply an textbf{Auxiliary Weight Network (AWN)} (indicated by $w_{f}$ in Fig. \ref{design} (c)) on the resulting feature maps after the layering sharing, since the feature maps of two network branches should carry different weights considering that they are extracted from different sensing modalities.

\begin{figure}[t]
\centering
            \includegraphics[width=0.8\linewidth]{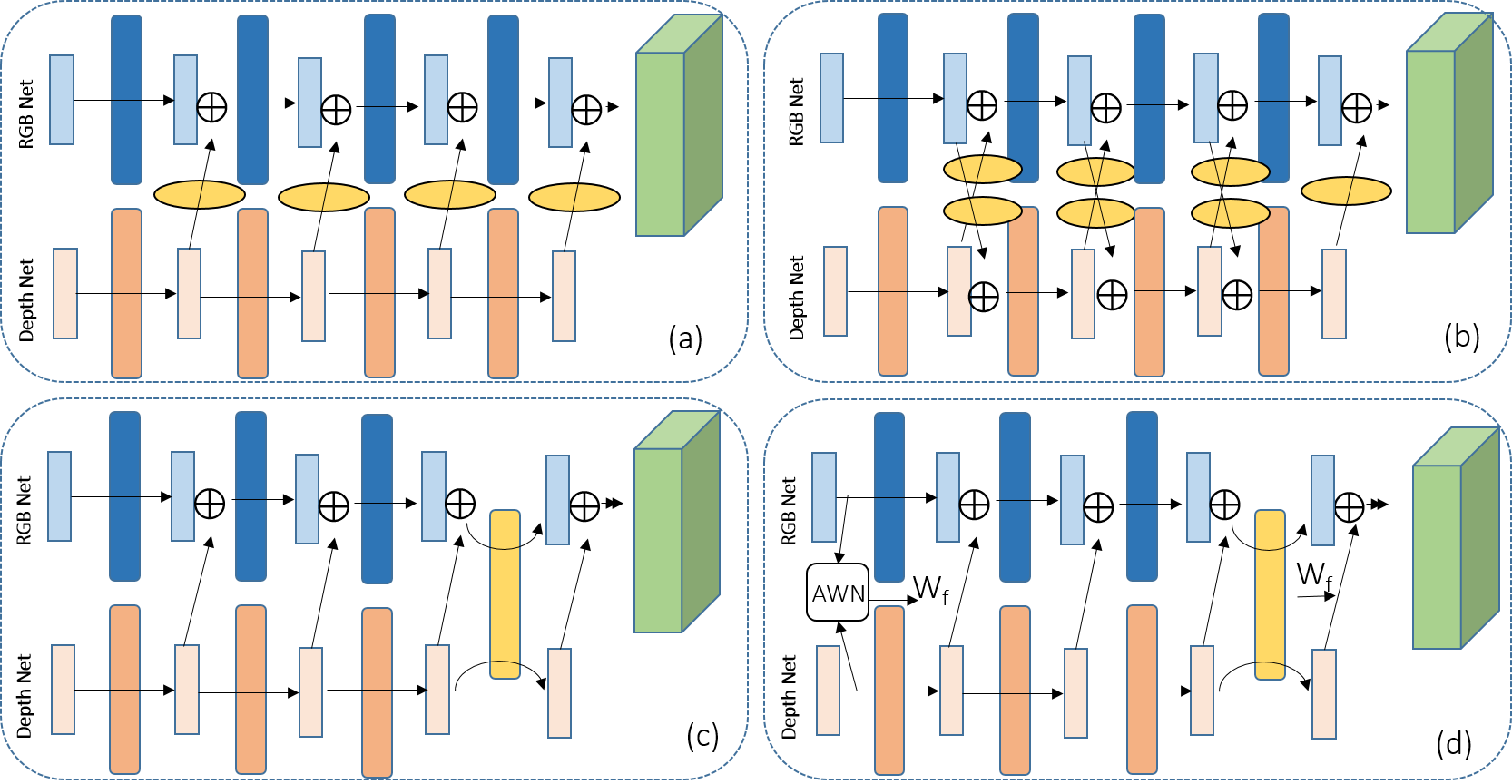}
            \vspace{0mm}
            \caption{Models with different fusion schemes: (a)AllFilter\_U, (b)AllFilter\_B, (c)BaseSharing, (d)WeightedSharing. }\label{combine}
            \vspace{-4mm}
\end{figure}

Specifically, we first adopt same conv layers to extract the high level feature maps from the two input sensing modalities. 
Then, the difference of the two sets of extracted high level feature maps will be fed into a stacked full-connected layer to generate the auxiliary weight parameter $W_{f}$, which represents the weight of feature maps in Depth Network branch when being fused with its counterpart RGB feature maps as depicted in Fig.\ref{combine} (d).

\begin{figure*}[t]
\centering
            \includegraphics[width=0.8\linewidth]{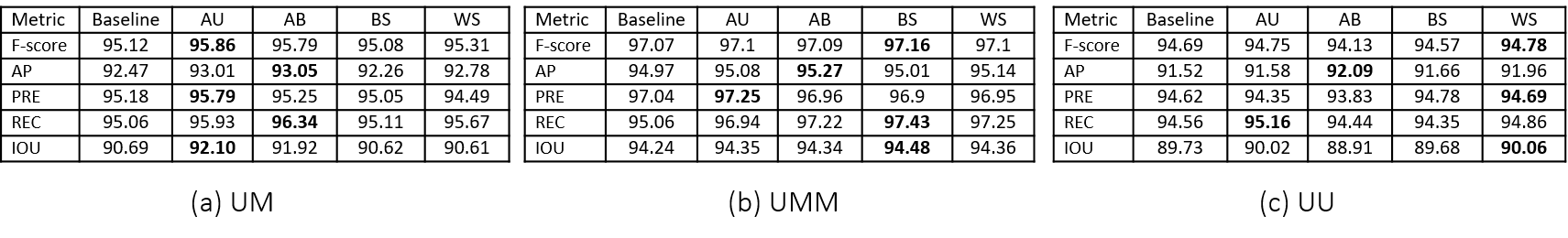}
            \vspace{-4mm} 
            \caption{Accuracy performance of proposed models under different road scenes in tables. In order to adapt the paper width, we term the architecture AllFilter\_U as AU, AllFilter\_B as AB, BaseSharing as BS and WeightedSharing as WS respectively. The best model is highlighted under each metric. }\label{radartable}
\end{figure*}

\subsection{Feature Disparity in Objective Loss} \label{fdloss}
Besides the objective loss function(i.e., segmentation loss), we further utilize the feature disparity metric as an additional sub-objective loss function (i.e., Feature Disparity Loss) to further constrain the model to gradually learn to extract features with the same characteristics yet complementary content from its own sensing modality.
Therefore, two modalities can together reach a consensus on what they are perceiving, a better resulting accuracy performance can be reaped, which will be demonstrated in Sec.\ref{eval}. 
With the Feature Disparity Loss included (indicated by the green box in Fig. \ref{design} (a) and (b)), the overall objective loss function can be formally described as follows,

\begin{equation}\label{final_loss}
\mathcal{L}_{loss}=\mathcal{L}_{Segmentation} + \alpha\sum_{i}\mathcal{D}_{fd-i}
\vspace{-2mm}
\end{equation}

Apart from segmentation loss $\mathcal{L}_{Segmentation}$, we formulate the Feature Disparity Loss $\mathcal{D}_{fd-i}$ at fusion stage $i$ by comparing the edge characteristic between the two sets of the feature maps instead of comparing feature maps directly. 
Moreover, a tuning knob $\alpha$ is assigned to the feature disparity loss to decide how much it weighs in the overall loss function.
Note that the proposed loss function will only be applied during the training process, thus it will not affect the inference latency.

\section{Evaluation}

\subsection{Experimental Setup}
\subsubsection{Training Environment}
The models proposed are trained and tested on a single NVIDIA’s Quadro RTX 8000 GPU with CUDA 10.0 and PyTorch 1.1. As discussed in \ref{basemodel}, we adopt RoadSeg \cite{sne} as our baseline, and the final objective loss is comprised of two parts as shown in Eq. \ref{final_loss}.
From our experimental experience, we empirically set $\alpha$ to 0.3.

\subsubsection{Dataset and Metrics}
KITTI road dataset \cite{kitti} is widely used for autonomous driving research as a benchmark. It contains 289 image pairs(RGB image and Depth image) for training and 290 image pairs for testing, where within each pair there both containing three different road scene categories including urban marked roads (UM), urban multiple marked lanes (UMM), and urban unmarked roads (UU). 
Note that the Depth images are generated from lidar point-cloud data by utilizing the pre-processing method proposed in the baseline \cite{sne}. Generally, there are four common metrics for performance evaluation: F-score, AP, PRE, REC, and IOU\cite{sne}.
For fair evaluation, KITTI does not provide ground-truths for testing images, and the final segmentation results of testing images would be converted to a bird's eye view before submitting to KITTI evaluation server.

\begin{figure}[t]
\centering
            \includegraphics[width=\linewidth]{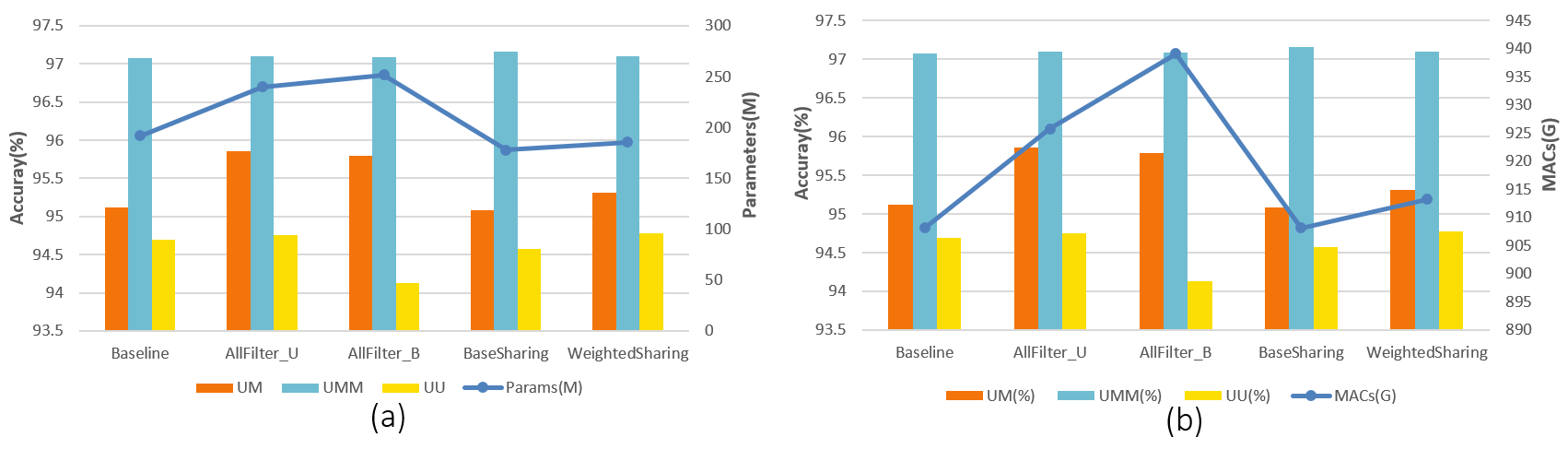}
            \vspace{-4mm}
            \caption{Model performance in terms of accuracy, total number of MACs and parameters. }\label{evalall}
            \vspace{-6mm}
\end{figure}



\subsection{Performance of Our Proposed Models}\label{eval}

With our proposed feature matching technique and layer-sharing method, various architectures can be derived from combining different fusion schemes with the baseline. The corresponding diagrammatic presentation can be found in Fig.\ref{combine}.  In Fig.\ref{combine} (a), we apply unidirectional Fusion-filter from Depth branch to RGB branch at each fusion stage and term this architecture as \textbf{AllFilter\_U}. Similarly, we call architecture from the Fig.\ref{combine} (b) as \textbf{AllFilter\_B} for applying bidirectional Fusion-filter. Fig.\ref{combine} (c) is termed as \textbf{BaseSharing} as the last convolutional stage been shared between two branches. \textbf{WeightedSharing} in figure (d) is named for the  Auxiliary Weight Network(AWN) applied on top of BaseSharing. Note that the yellow oval box represents Fusion-filters and the yellow square box represents the shared layer, and the Auxiliary Weight Network is presented in a white box in Fig.\ref{combine} (d). 
We experiment with these models first by training them on the GPU platform and then evaluate the trained model with testing images. Note that for the baseline model, we can achieve as best as 95.12\% for UM, 97.07\% for UMM, and 94.69\% for UU in our experimental environment respectively, which is lower than reported in \cite{sne}.

\subsubsection{Accuracy and Corresponding Computational Overhead} \label{comparison} 

As we can see in tables of Fig. \ref{radartable}, the overall accuracy performance of all metrics are presented under three evaluation road scenes. Generally, our proposed models perform better than baseline nearly in every metric under different road scenes. However, the accuracy improvement of our proposed models under UU is less significant, since UU is the most challenging one. Among all metrics,  AllFilter\_U performs better than others in UM, the same as BaseSharing in UMM, and WeightedSharing in UU category while they all top three metrics out of five as shown in tables in Fig. \ref{radartable}.

For models with the Fusion-filter technique, we have evaluated two types of models, \textbf{AllFilter\_U} and \textbf{AllFilter\_B}. By taking advantage of the unidirectional fusion-filter, AllFilter\_U is able to carry out fusion with more matched feature maps, which is witnessed as the orange line shown in Fig.\ref{badmatch} (a). It outperforms baseline in every category and tops three accuracy metrics out of five in UM category, one in UM and one in UU. With bidirectional Fusion-filter, AllFilter\_B outperforms baseline in UM and UMM category in nearly every metric. The reason might be with the help of Fusion-filters across two branches, more balanced segmentation results can be obtained. However, with the introduction of Fusion-filters, higher computational overhead in terms of MACs and parameters is introduced as we can see in Fig. \ref{evalall}.

For the Layer-sharing method, both \textbf{BaseSharing} and \textbf{WeightedSharing} are evaluated.  BaseSharing achieves three best accuracy performance out of five metrics in UMM road scene category with least computational cost as observed from Fig. \ref{radartable} and Fig. \ref{evalall}.
And after AWN is applied on top of BaseSharing, WeightedSharing outperforms Baseline in all three road scenarios and all metrics, and it even achieves the best performance under three metrics in the challenging UU scenario among all the proposed models as the introduced weight parameter can dynamically adjust the weight from one network branch to the other based on different input during fusion. Moreover, WeightedSharing still carries less model parameters than Baseline as shown in Fig.\ref{evalall}.

\begin{figure}[t]
\centering
            \includegraphics[width=0.8\linewidth]{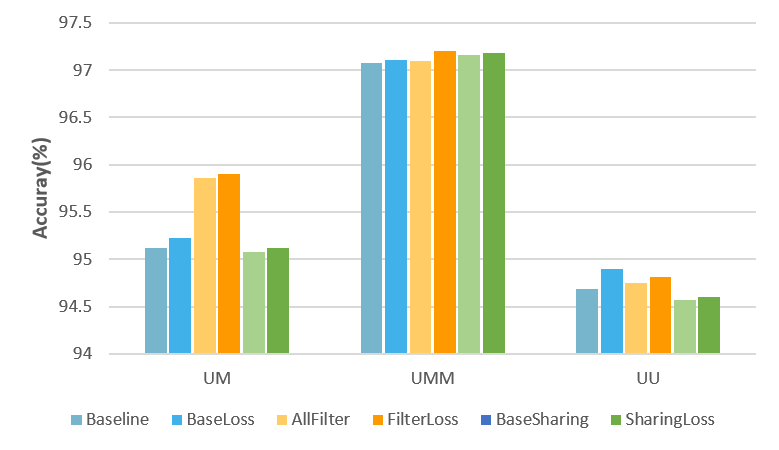}
            \vspace{-3mm}
            \caption{Ablation study for Feature Disparity Loss. 
            The "loss" indicates the Feature Disparity Loss plus segmentation loss.
            }\label{chart}
\end{figure}

\subsubsection{Ablation Study for Feature Disparity Loss}
We conduct ablation evaluation on our proposed models with Feature Disparity Loss, as shown in Fig. \ref{chart}. Three typical architectures (Baseline,  AllFilter\_U and BaseSharing) are evaluated using the same KITTI dataset. Specifically, Baseline,  AllFilter\_U and BaseSharing are trained with only segmentation loss while  BaseLoss,  FilterLoss and SharingLoss are trained with both segmentation loss and the proposed Feature Disparity loss.
As shown in the figure, architectures trained with Feature Disparity Loss outperforms the same architecture trained without it. This result proofs the efficacy of the concept of feature-matching. Especially, for Baseline trained with Feature Disparity Loss, BaseLoss even achieves the best accuracy in UU road scenario under F-score metric.

\subsubsection{Qualitative Samples}
\begin{figure}[t]
\centering
            \includegraphics[width=0.8\linewidth]{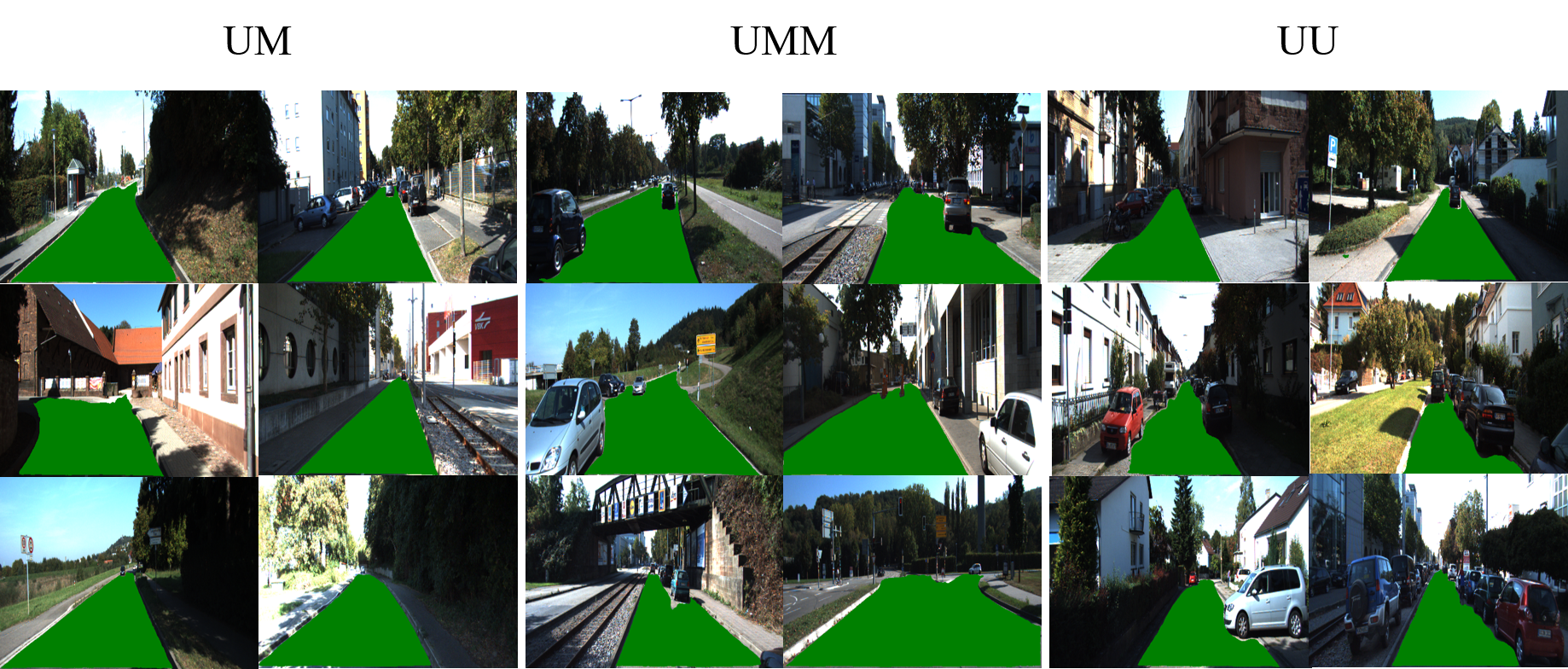}
            \vspace{-3mm}
            \caption{Example results under different road scenes.  }\label{demo}
            \vspace{-4mm}
\end{figure}

Fig. \ref{demo} shows some qualitative results of our proposed model-AllFilter on the test set of the KITTI benchmark. Three road scenes(UM, UMM, UU) are presented from left to right. To prove the robustness of our model, we deliberately select the road scenes with over-exposure or shadows. From these samples, we can see that our model is robust to these adversarial lighting conditions on the road.

\section*{Conclusion}

In order to achieve better perception accuracy, we investigate the various DCNN fusion architectures, and we identify that feature-mismatching issue exists in directly element-wisely adding feature maps from different sensing modalities, which is commonly adopted by most DCNN middle fusion methods. To tackle this issue, we first propose a feature disparity metric that quantitatively measures the degree of feature disparity between feature maps to be fused. After learning that the feature-mismatching issue exists in conventional element-wise middle fusion architecture, we propose a feature-matching technique, Fusion-filter, to address the issue. We further learn that feature maps in deeper layer appears to possess less different features, the Layer-sharing approach is thus proposed to reduce the model's computational overhead. And together with the Feature Disparity Loss, proposed models can learn corresponding features from different modalities to achieve better accuracy. 
Experimental results demonstrate that our proposed feature matching techniques can achieve better accuracy than baseline while the Layer-sharing achieves comparable accuracy with less computational overhead compared to the baseline.

\vspace{12pt}


\begin{thebibliography}{00}

\bibitem{cross} Caltagirone, L., Bellone, M., Svensson, L. and Wahde, M., 2019. LIDAR–camera fusion for road detection using fully convolutional neural networks. Robotics and Autonomous Systems, 111, pp.125-131.

\bibitem{progressive} Chen, Z., Zhang, J. and Tao, D., 2019. Progressive lidar adaptation for road detection. IEEE/CAA Journal of Automatica Sinica, 6(3), pp.693-702.



\bibitem{sne} Fan, R., Wang, H., Cai, P. and Liu, M., 2020, August. Sne-roadseg: Incorporating surface normal information into semantic segmentation for accurate freespace detection. In European Conference on Computer Vision (pp. 340-356). Springer, Cham.

\bibitem{fusenet} Hazirbas, C., Ma, L., Domokos, C. and Cremers, D., 2016, November. Fusenet: Incorporating depth into semantic segmentation via fusion-based cnn architecture. In Asian conference on computer vision (pp. 213-228). Springer, Cham.



\bibitem{lecun} LeCun, Y., Kavukcuoglu, K. and Farabet, C., 2010, May. Convolutional networks and applications in vision. In Proceedings of 2010 IEEE international symposium on circuits and systems (pp. 253-256). IEEE.

\bibitem{kitti} Geiger, A., Lenz, P. and Urtasun, R., 2012, June. Are we ready for autonomous driving? the kitti vision benchmark suite. In 2012 IEEE Conference on Computer Vision and Pattern Recognition (pp. 3354-3361). IEEE.

\bibitem{early1} Wulff, F., Schäufele, B., Sawade, O., Becker, D., Henke, B. and Radusch, I., 2018, June. Early fusion of camera and lidar for robust road detection based on U-Net FCN. In 2018 IEEE Intelligent Vehicles Symposium (IV) (pp. 1426-1431). IEEE.


\bibitem{late1} Du, X., Ang, M.H., Karaman, S. and Rus, D., 2018, May. A general pipeline for 3d detection of vehicles. In 2018 IEEE International Conference on Robotics and Automation (ICRA) (pp. 3194-3200). IEEE.


\bibitem{filterslide} Dumoulin, V. and Visin, F., 2016. A guide to convolution arithmetic for deep learning. arXiv preprint arXiv:1603.07285.

\bibitem{opencvedge} Xie, G. and Lu, W., 2013. Image edge detection based on opencv. International Journal of Electronics and Electrical Engineering, 1(2), pp.104-106.

\bibitem{yolo} Redmon, J., Divvala, S., Girshick, R. and Farhadi, A., 2016. You only look once: Unified, real-time object detection. In Proceedings of the IEEE conference on computer vision and pattern recognition (pp. 779-788).

\bibitem{imgtask1} Krizhevsky, A., Sutskever, I. and Hinton, G.E., 2012. Imagenet classification with deep convolutional neural networks. Advances in neural information processing systems, 25, pp.1097-1105.

\bibitem{imgtask3} Simonyan, K. and Zisserman, A., 2014. Very deep convolutional networks for large-scale image recognition. arXiv preprint arXiv:1409.1556.


\bibitem{resnet} He, K., Zhang, X., Ren, S. and Sun, J., 2016. Deep residual learning for image recognition. In Proceedings of the IEEE conference on computer vision and pattern recognition (pp. 770-778).


\bibitem{survey} Haghighat, M.B.A., Aghagolzadeh, A. and Seyedarabi, H., 2011. A non-reference image fusion metric based on mutual information of image features. Computers \& Electrical Engineering, 37(5), pp.744-756.


\bibitem{ssim} Wang, Z., Bovik, A.C., Sheikh, H.R. and Simoncelli, E.P., 2004. Image quality assessment: from error visibility to structural similarity. IEEE transactions on image processing, 13(4), pp.600-612.

\bibitem{mi} Qu, G., Zhang, D. and Yan, P., 2002. Information measure for performance of image fusion. Electronics letters, 38(7), pp.313-315.

\bibitem{crossbin} Ling, H. and Okada, K., 2006, June. Diffusion distance for histogram comparison. In 2006 IEEE Computer Society Conference on Computer Vision and Pattern Recognition (CVPR'06) (Vol. 1, pp. 246-253). IEEE.

\bibitem{basu2002gaussian}  Basu, M., 2002. Gaussian-based edge-detection methods-a survey. IEEE Transactions on Systems, Man, and Cybernetics, Part C (Applications and Reviews), 32(3), pp.252-260.

\end{thebibliography}
\end{document}